\documentclass{article}
\usepackage{color}

\usepackage{ijcai24}

\usepackage{times}
\usepackage{soul}
\usepackage{bm}
\usepackage{url}
\usepackage[hidelinks]{hyperref}
\usepackage[utf8]{inputenc}
\usepackage[small]{caption}
\usepackage{graphicx}
\usepackage{amsmath}
\usepackage{amsthm}
\usepackage{bbm}
\usepackage{booktabs}
\usepackage{algorithm}
\usepackage{algorithmic}
\usepackage[switch]{lineno}
\usepackage{afterpage}
\usepackage{stfloats}
\usepackage{multirow}
\usepackage{hhline}
\usepackage{array}

\title{Negative-Binomial Randomized Gamma Markov Processes for Heterogeneous Overdispersed Count Time Series}

\author{
Rui Huang$^1$
\and
Sikun Yang$^1$\and
Heinz Koeppl$^{2}$
\affiliations
$^1$School of Computing and Information Technology, Great Bay University\\
$^2$Department of Electrical Engineering and Information Technology, Technische Universit\"at Darmstadt
}

\begin{document}

\maketitle

\begin{abstract}
Modeling count-valued time series has been receiving increasing attention since count time series naturally arise in physical and social domains. Poisson gamma dynamical systems (PGDSs) are newly-developed methods, which can well capture the expressive latent transition structure and \emph{bursty} dynamics behind count sequences. In particular, PGDSs demonstrate superior performance in terms of data imputation and prediction, compared with canonical linear dynamical system (LDS) based methods. Despite these advantages, PGDS cannot capture the \emph{heterogeneous} overdispersed behaviours of the underlying dynamic processes. To mitigate this defect, we propose a negative-binomial-randomized gamma Markov process, which not only significantly improves the predictive performance of the proposed dynamical system, but also facilitates the fast convergence of the inference algorithm. Moreover, we develop methods to estimate both factor-structured and graph-structured transition dynamics, which enable us to infer more explainable latent structure, compared with PGDSs. Finally, we demonstrate the explainable latent structure learned by the proposed method, and show its superior performance in imputing missing data and forecasting future observations, compared with the related models.
\end{abstract}

\section{Introduction}
Count time sequences, naturally arise in many domains such as text mining ~\cite{blei2006dynamic,Wang2008ContinuousTD,BleiWWW18,10.1145/3219819.3219995,Dieng2019TheDE}, cell genomic analysis~\cite{levitin2019novo,TrajNet,GPSPTIAL}, population movement forecasting ~\cite{pmlr-v28-sheldon13,IOT,roy2020nonparametric}, and etc. Modeling count sequences has been drawing increasing research attention because these real-world count data usually exhibit \emph{bursty} and \emph{overdispersed} behaviours, which cannot be well-captured by canonical linear dynamical systems (LDSs)~\cite{ghahramani1998learning}. In addition, some previous works  use extended rank likelihood functions ~\cite{han2014dynamic} which link the count observations to latent continuous dynamics to model count time sequences. Nonetheless, the extended rank likelihood functions cannot faithfully capture \emph{bursty} dynamics underlying real-world count sequences.  Meanwhile, the extended rank likelihood functions often require an approximate inference scheme, and thus scale poorly with high-dimensional count sequences, such as single-cell RNA sequencing data~\cite{Dunson_scRNAseq}. 
Notably, some recent works~\cite{acharya2015nonparametric,schein2016poisson,pmlr-v48-schein16,schein2019poisson} model sequential count observations using gamma Poisson family distributions. More specifically, \cite{acharya2015nonparametric} develops a gamma Markov process to capture continuous dynamics underlying count-valued sequences. In particular, the number of latent factors behind high-dimensional count data, can be appropriately determined by the gamma process prior, in a Bayesian non-parametric manner. Following the success of \cite{acharya2015nonparametric}, \citeauthor{schein2016poisson}~\shortcite{schein2016poisson} study a Poisson gamma dynamical system, in which a transition kernel is designed to capture how the latent dimensions interact with each other to model complicated observed dynamics. Another appealing aspect of the Poisson gamma dynamic model is that the posterior simulation can be performed using a tractable-yet-efficient Gibbs sampling algorithm via Poisson-Logarithm data augmentation strategy~\cite{ZhouAug,10.1109/TPAMI.2013.211}. Hence, the Poisson gamma dynamic models~\cite{acharya2015nonparametric,schein2016poisson,schein2019poisson} are in particular well-fitted to impute missing entries, to predict future unseen observations and to estimate uncertainties. 

Despite these advantages, these models still cannot well capture the \emph{heterogeneous} overdispersion effects of the latent dynamic processes behind count observations. 
For instance, international event data, usually consists of multiple latent dynamic processes, which often change rapidly with the different magnitudes~\cite{InternationalRelationsData,Stewart2014LatentFR}. 
To capture such \emph{heterogeneous} overdispersed behaviours, we develop a negative-binomial-randomized gamma Markov chain structure,  which not only greatly enhances the model flexibility, but also facilitates the fast convergence of the derived Gibbs sampling algorithms. Moreover, the transition dynamics behind real-world high-dimensional count data, are often \emph{sparse}, and exhibit a certain amount of graph structure. Hence, we propose to learn the graph-structured transition dynamics using relational gamma processes~\cite{EPM}. To the best of our knowledge, this is the first attempt to learn the latent graph-structured transition dynamics under the Poisson gamma dynamical system.

The main contributions of the paper include: 1) A negative-binomial-randomized gamma Markov process (NBRGMP) is proposed to estimate the \emph{heterogeneous} overdispersion effects of the latent dimensions underlying sequential count observations; 2) Relational gamma processes are thoroughly studied to learn both factor-structured and graph-structured transition dynamics, which renders the estimated latent structure more explainable, compared with transition structure inferred using non-informative priors; 3) Although the proposed NBRGMP and its factor-structured and graph-structured extensions are intractable, simple-yet-efficient Gibbs sampling algorithms are developed via Negative-binomial data augmentation strategies to perform inference; 4)Extensive experiments are conducted to illustrate the explainable transition structure learned by the proposed model. We demonstrate the superior performance of the proposed method in missing data imputation and future snapshot forecasting, with several related works.

\section{Preliminary}
Suppose we have a sequentially-observed count data over time interval $[0,T]$ specified by $\mathbf{N}=(\mathbf{n}_1, \ldots, \mathbf{n}_V )^{\mathrm T}$ of V dimensions, where $\mathbf{n}_v = ({n}_v^{(1)},\ldots,{n}_v^{(T)})^{\mathrm T}$ with $n_{v}^{(t)}$ denoting the $v$-th observation at time $t$.  
The Poisson gamma dynamical system~\cite{schein2016poisson} models the count $n_v^{(t)}$ as 
\begin{align}
    \label{eq:1.1}
    n_v^{(t)} \sim \mathrm{Pois}(\delta^{(t)}\sum\nolimits_{k=1}^{K}\phi_{vk}\theta_k^{(t)}),
\end{align}
where $\theta_k^{(t)}$ captures the strength of latent component $k$ at time $t$, and $\phi_{vk}$ represents the involvement degree of dimension $v$ in latent component $k$. To model the underlying dynamics, the PGDS assumes that the latent components evolve over time according to a gamma Markov chain structure as
\begin{align}
    \label{eq:1.2}
    \theta_k^{(t)} \sim \mathrm{Gam}(\tau_0 \sum\nolimits_{k_2=1}^{K} \pi_{kk_2}      \theta_{k_2}^{(t-1)}, \tau_0 ),
\end{align}
where the latent components $\bm{\theta}^{(t-1)}=(\theta^{(t-1)}_1, \ldots, \theta^{(t-1)}_K)^{\mathrm{T}}$ evolve over time through the transition matrix $\bm{\Pi}$. The $\theta^{(t-1)}_k$ captures how strongly the $k$-th latent component activates at time $t-1$, and $\pi_{kk_2}$ models how strongly the $k_2$-th component $\theta_{k_2}^{(t-1)}$ at time $t-1$ affect the $k$-th component $\theta_k^{(t)}$ at time $t$. Eq.\ref{eq:1.2} naturally defines a gamma Markov chain structure. The expectation and variance of the gamma Markov chain can be calculated respectively as $\mathsf{E}[\bm{\theta}^{(t)} \mid \bm{\theta}^{(t-1)}, \bm{\Pi}] = \bm{\Pi} \bm{\theta}^{(t-1)}$ and $\mathsf{Var}[\bm{\theta}^{(t)} \mid \bm{\theta}^{(t)}, \bm{\Pi}] = (\bm{\Pi} \bm{\theta}^{(t-1)}) \tau_0^{-1}$, where $\tau_0$ controls the variance of $\bm{\theta}^{(t)}$.

\citeauthor{schein2019poisson}~\shortcite{schein2019poisson} further develop a Poisson-randomized gamma Markov chain (PRGMC) structure specified by
\begin{equation}\theta_k^{(t)} \sim \mathrm{Gam}(\epsilon_0^{(\theta)}+h_k^{(t)}, \tau), \ \ h_k^{(t)} \sim \mathrm{Pois}(\tau \sum\limits_{k_2} \pi_{kk_2} \theta_{k_2}^{(t-1)}).\notag\end{equation}

By marginalizing out the Poisson latent states $h_k^{(t)}$, we have a continuous-valued dynamical system given by 
\begin{equation}
\theta_k^{(t)} \sim \mathrm{RG1}(\epsilon_0^{(\theta)}, \tau\sum\nolimits_{k_2}\pi_{kk_2}\theta_{k_2}^{(t-1)}, \tau),\notag
\end{equation}
where RG1 is the randomized gamma distribution of the first type. 
The marginal expectation and variance of the PRGMC is $\mathsf{E}[\bm{\theta}^{(t)} \mid \bm{\theta}^{(t-1)}, \bm{\Pi}] = \bm{\Pi} \bm{\theta}^{(t-1)}+\epsilon_0^{(\theta)}\tau^{-1}$ and \\
$\mathsf{Var}[\bm{\theta}^{(t)} \mid \bm{\theta}^{(t-1)},\bm{\Pi}]=2\bm{\Pi \theta^{(t-1)}}\tau^{-1} + \epsilon_0^{(\theta)}\tau^{-2}$, respectively.

\section{The Proposed Model}
In this section we will introduce the novel negative-binomial-randomized gamma Markov chain structure to capture the \emph{heterogeneous} overdispersion effects of the latent dimensions behind count data. Then we shall describe how to learn explainable latent transition structure with relational gamma processes. The proposed 
negative-binomial-randomized gamma 
dynamical system is defined by 
\begin{equation}
    \label{eq:10}
    n^{(t)}_v \sim \mathrm{Pois}(\delta^{(t)} \sum\nolimits_{k=1}^{K} \lambda_k \phi_{vk} \theta^{(t)}_{k}),
\end{equation}
where $\delta^{(t)}$ is a nonnegative multiplicative term capturing time-dependent bursty dynamics. We place a gamma prior on $\delta^{(t)}$ as $\delta^{(t)} \sim \mathrm{Gam}(\epsilon_0, \epsilon_0)$, and let $\delta^{(t)}=\delta$ if the generative  process (Eq. \ref{eq:10}) is stationary over time. Here $\bm{\phi}_k = (\phi_{1k}, \phi_{2k}, \ldots, \phi_{Vk})^\mathrm{T}$ denotes the loading coefficient of $k$-th latent component, and $\lambda_k$ denotes the weight of $k$-th latent component. To ensure model identifiability, we require $\sum_v\phi_{vk}=1$ and thus have a Dirichlet prior over $\bm{\phi}_k$ given by $\bm{\phi}_{k} \sim \mathrm{Dir}(\epsilon_0,...,\epsilon_0)$.  
More specifically, we draw $\lambda_k$ from a hierarchical prior as $\lambda_k \sim \mathrm{Gam}(\frac{\epsilon_0^{(\lambda)}}{K}+g_k, \beta)$, in which  $g_k \sim \mathrm{Pois}(\frac{\gamma}{K})$. We specify gamma priors over $\gamma$ and $\beta$ as $\gamma \sim \mathrm{Gam}(\epsilon_0, \epsilon_0), \beta \sim \mathrm{Gam}(\epsilon_0, \epsilon_0)$. Note that as $K \to \infty$, the summation of the weight expectation remains finite and fixed, i.e., $\sum_{k=1}^{\infty}\mathsf{E}[\lambda_k] = \beta^{-1}(\epsilon_0^{(\lambda)} + \gamma)$. Hence, this hierarchical prior enables us to effectively estimate a finite number of latent factors that are representative to capture the temporal dynamics.

\subsection{Negative-Binomial Randomized Gamma Markov Processes.}
\begin{figure}[t]
    \centering
    \includegraphics[scale = 0.43]{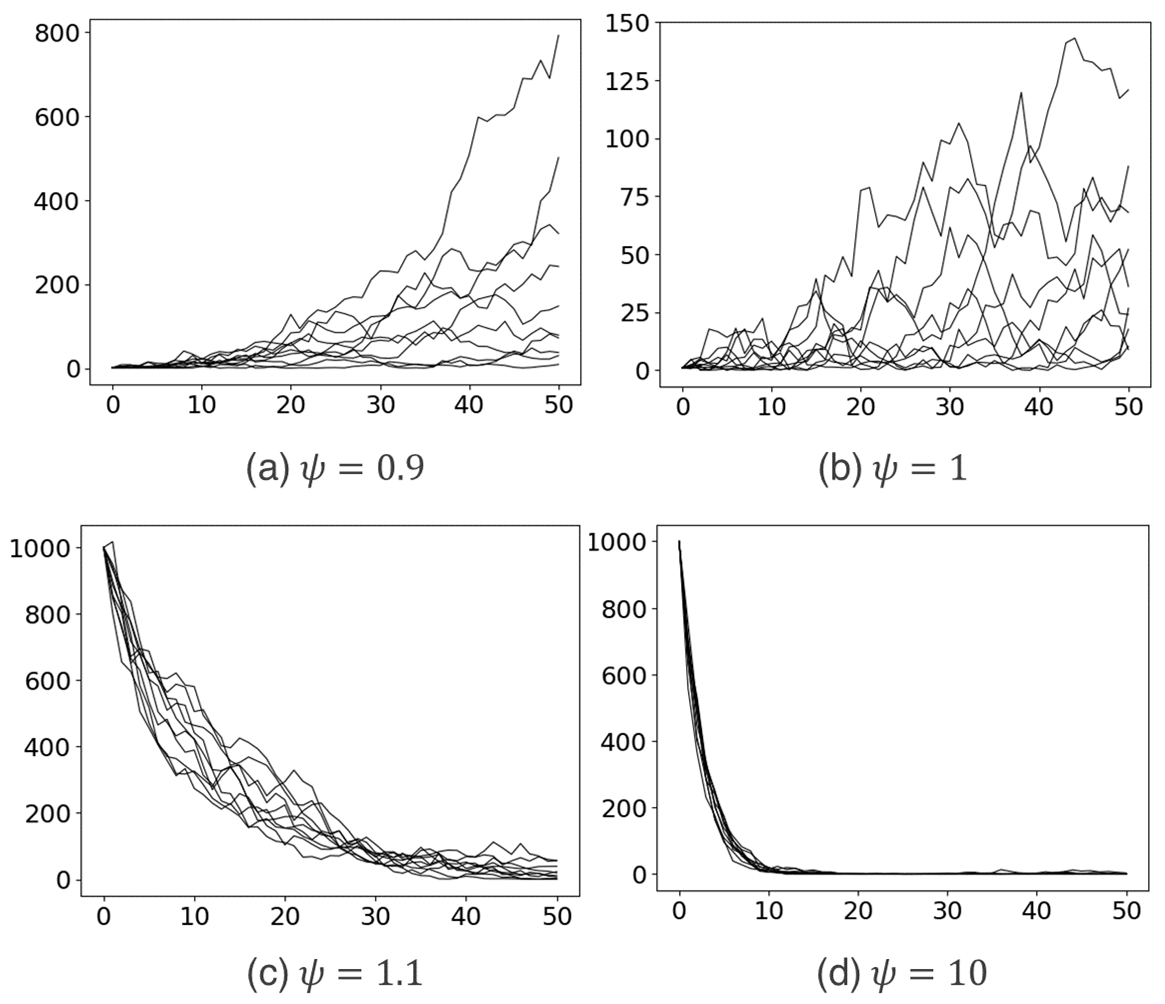}
    \caption{The realizations of the negative-binomial-randomized gamma Markov processes defined in Eq.\ref{eq:13}. Here $\epsilon_0^{(\theta)}$ and $\tau$ were set to 0 and 1, respectively. The initial values of the NBRGMPs in (a) and (b), were set to 1, (c) and (d) were set to 1000, and the chains were simulated until $t=50$. Each subplot contains ten independent realizations.}
    \label{fig:realization}
\end{figure}

To capture the \emph{heterogeneous} overdispersed behaviors of the latent dimensions behind count sequences, we introduce a negative-binomial randomized gamma Markov process (NBRGMP) specified by
\begin{equation}
\begin{aligned}
    \label{eq:13}
    \theta^{(t)}_{k} &\sim \mathrm{Gam}(\epsilon^{(\theta)}_0 + h^{(t)}_{k}, \tau), \\
    h_k^{(t)} &\sim \mathrm{NB}(\tau \sum\nolimits_{k_2=1}^{K}\pi_{kk_2} \theta_{k_2}^{(t-1)}, \frac{\psi}{1+\psi}),
\end{aligned}
\end{equation}
where we set $\theta_{k}^{(0)} = \lambda_k$, and $\theta_k^{(t)}$ is gamma distributed with shape parameter $\epsilon_0^{(\theta)} + h_k^{(t)}$ where $\epsilon_0^{(\theta)} \geq 0$, and the rate parameter $\tau$. Here, instead of specifying a Poisson prior as did in ~\cite{schein2019poisson}, we draw the intermediate latent state $h_k^{(t)}$ from a negative-binomial distribution to enhance the flexibility of $\theta_k^{(t)}$, which enable us to estimate the heterogeneous overdispersed behaviours of the latent dynamic processes. 

More specifically, the marginal expectation and variance of the NBRGMP can be calculated by iteration as
\begin{align}
\mathsf{E}[\bm{\theta}^{(t)} \mid \bm{\theta}^{(t-1)}] &= \epsilon_0^{(\theta)} \tau^{-1} + \frac{\bm{\Pi} \bm{\theta}^{(t-1)}}{\psi},\notag\\
\mathsf{Var}[\bm{\theta}^{(t)} \mid \bm{\theta}^{(t-1)}] &= \epsilon_0^{(\theta)} \tau^{-2} + \frac{(1+2\psi)\bm{\Pi}\bm{\theta}^{(t-1)}}{\psi^2 \tau},\notag
\end{align}
respectively. 
We note that both the concentration parameter $\tau$ and hyperparameter $\epsilon_0^{(\theta)}$ appear in the additive term of the expectation and variance, which can be concealed by letting $\epsilon_0^{(\theta)}=0$. The hyperparameter $\psi$ plays a crucial role in controlling the variance of the NBRGMP. More specifically, when $\psi \in (0, 1)$ the values of $\bm{\theta}^{(t)}$ will fluctuate dramatically because of its large expectation and variance. When $\psi=1$, the expectation of the NBRGMP will be the same with the PRGMC (as discussed in Sec.2), while the variance of the NBRGMP will be three times of the variance of the PRGMC, which thus allows the proposed dynamical system to capture reasonable heterogeneous overdispersed behaviors. When $\psi \in (1, 1+\sqrt{2})$, the expectation of the NBRGMP will tend to be smaller compared with the expectation of the PRGMC, which will be more suitable to capture sparse counts. Meanwhile, the variance of the NBRGMP still allows us to capture a limited range of overdispersion effects. If $\psi \ge 1+\sqrt{2}$, both expectation and variance converge to zeros as $\psi$ goes to infinity. Fig.~\ref{fig:realization} plots the realizations of the NBRGMP by varying the parameter $\psi$. 
Note that the negative-binomial distributed latent state $h_k^{(t)}$ can be equivalently drawn from a gamma-Poisson mixture as 
\begin{align}
h_k^{(t)} \sim \mathrm{Pois}(\hat{h}_k^{(t)}),\ \ \hat{h}_k^{(t)} \sim \mathrm{Gam}(\tau \sum\nolimits_{k_2} \pi_{kk_2} \theta_{k_2}^{(t-1)}, \psi).\notag\end{align}

\begin{figure}[t]
    \centering
    \includegraphics[scale = 0.385]{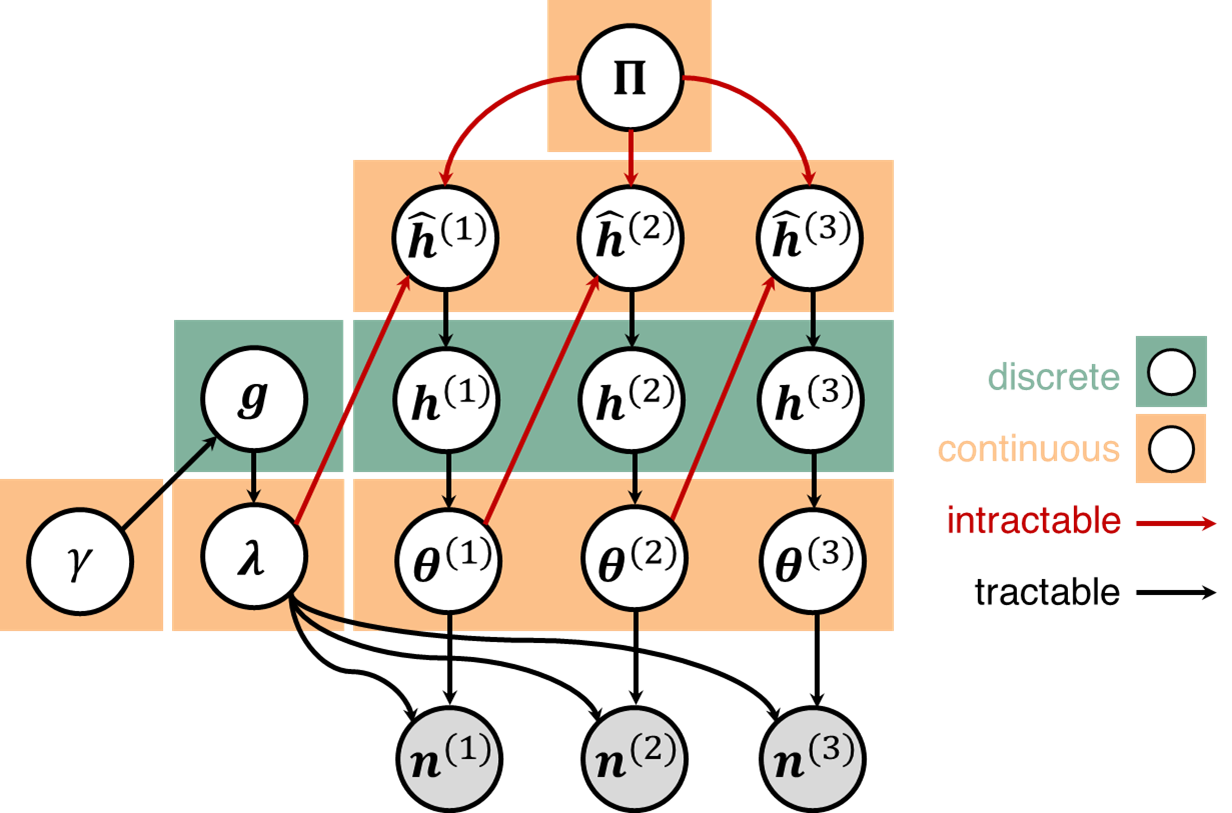}
    \caption{The hierarchical structure of the NBRGMP. The red arrows indicate intractable dependencies that require data augmentation schemes for posterior inference.}
    \label{fig:1}
\end{figure}

Fig.~\ref{fig:1} shows the graphical representation of the developed NBRGMP. When $\tau\sum_{k_2}\pi_{kk_2}\theta_{k_2}^{(t-1)} \to \infty$, the $h_k^{(t)}$ is approximately characterized by $\mathrm{Pois}(\frac{\tau}{\psi}\sum\nolimits_{k_2}\pi_{kk_2}\theta_{k_2}^{(t-1)})$. Hence, by marginalizing the Poisson distributed latent states $h_k^{(t)}$ from Eq.\ref{eq:13}, the negative-binomial randomized gamma dynamical system can be equivalently represented by randomized gamma distribution of the first type as 
\begin{align}
\theta_k^{(t)} \sim \mathrm{RG1}(\epsilon_0^{(\theta)}, \frac{\tau}{\psi}\sum\nolimits_{k_2}\pi_{kk_2}\theta_{k_2}^{(t-1)}, \tau).\notag\end{align} 

\subsection{Factor-structured Transition Dynamics} 
We first propose to learn the latent factor structure behind transition dynamics. To that end, we specify a hierarchical Dirichlet prior over $\bm{\pi}_k$ as
$\bm{\pi}_{k}\sim \mathrm{Dir}(a_{1k},\ldots, a_{Kk})$,
where $\mathbf{a}_k = (a_{1k},\ldots, a_{Kk})^{\mathrm T}$ is the hyper-parameter. Our goal here is to capture the correlation structure between the latent dimensions of the transition kernel. Thus, we model the hyper-parameter $\mathbf{A} = [a_{k_1k_2}]_{k_1,k_2}^K$ using a Poisson factor model as
\begin{align}
    a_{k_1k_2} \sim \mathrm{Pois}(\sum\nolimits_{c=1}^{C} m_{k_1c} r_c m_{k_2c}) \notag
,\end{align}
where $r_c$ is the weight of $c$-th latent factor, and $m_{kc}$ captures how strongly $k$-th component associate with $c$-th factor. Naturally, $k_1$-th component interact with $k_2$-th component through the weight $\sum\nolimits_{c=1}^{C} m_{k_1c} r_c m_{k_2c}$. To ensure the latent factor to be nonnegative, we draw the factor $r_c$, and factor loading $m_{kc}$ from the priors specified by
\begin{align}
    m_{kc} \sim \mathrm{Gam}(\hat{a}_k, \hat{b}_k),\ \ r_c \sim \mathrm{Gam}(\frac{r_0}{C}, c_0),\notag
\end{align}
respectively. Here, $C$ is the maximum number of latent factors. As $C\rightarrow\infty$, the weights of the latent factors $\{r_c\}_c^C$ and the factor loading $\{\mathbf{m}_{c}\}_c^C$ can be considered as a draw $G=\sum_{c=1}^{\infty}r_c\delta_{\mathbf{m}_c}$ from a gamma process $\mathrm{GaP}(G_0, c_0)$, where $G_0$ denotes the base measure over the metric space $\Omega$, and $c_0$ the concentration parameter~\cite{BNP}.

\subsection{Graph-Structured Transition Dynamics}
\begin{figure}[t]
    \centering
    \includegraphics[scale=0.4]{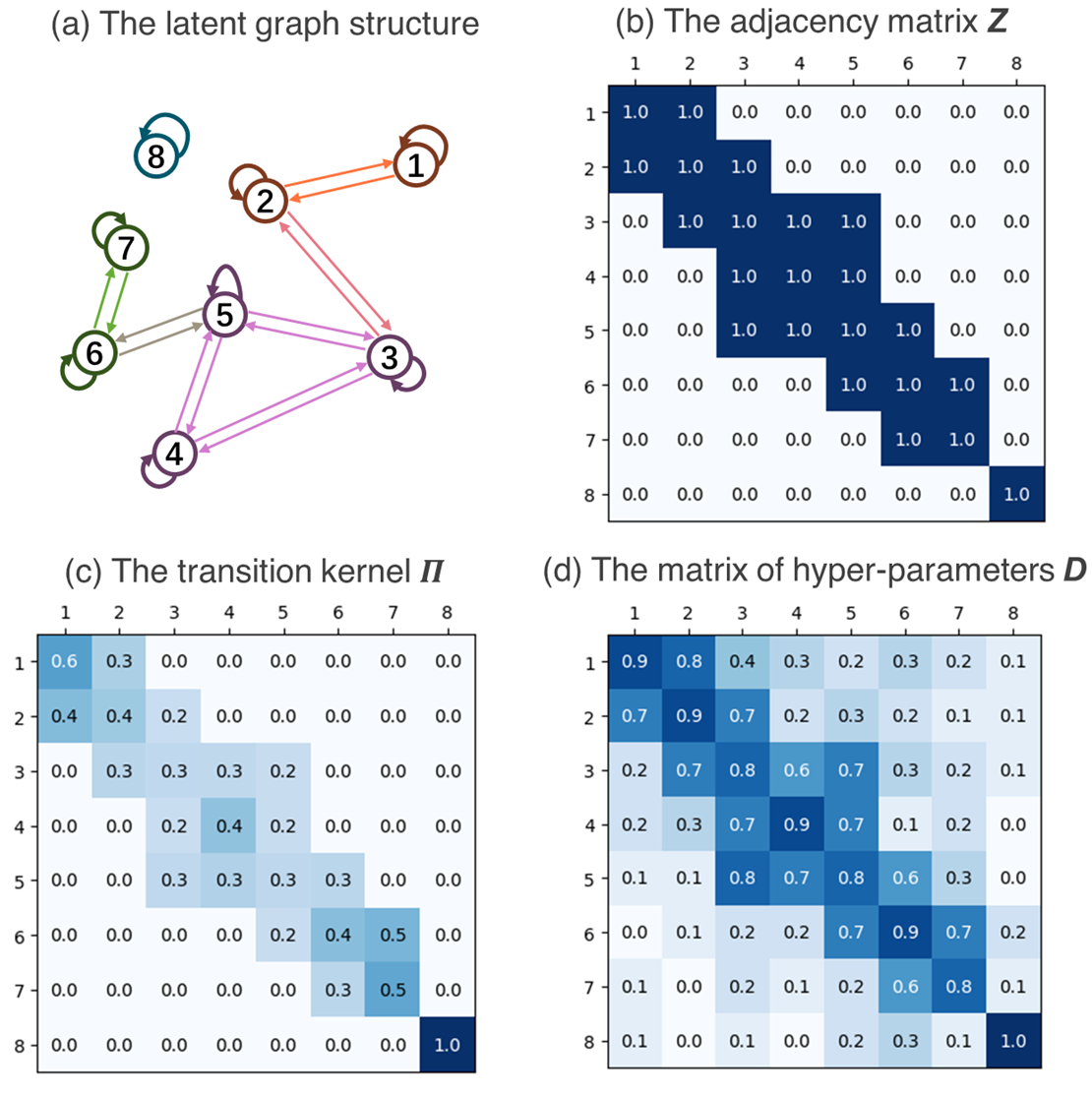}
    \caption{The graph structure of the latent dimensions of the transition kernel behind sequential count observations.}
    \label{fig:2}
\end{figure}

For high-dimensional count sequences, the underlying transition dynamics are often \emph{sparse} and exhibit a certain amount of graph structure. Hence, we further study to learn the latent graph-structured transition kernel behind count time series, using relational gamma process prior. In particular,  we sample the transition parameter $\bm{\pi}_k$ from a hierarchical Dirichlet prior, as $\bm{\pi}_{k} \sim \mathrm{Dir}(a_{1k},\ldots, a_{Kk})$. To introduce a sparse graph-structured transition kernel, we model the matrix of the hyper-parameter $\mathbf{A}=[a_{k_1k_2}]_{k_1,k_2}^K$ as
$\mathbf{A} = \mathbf{D}\odot\mathbf{Z},$
where 
$\mathbf{D} = [d_{k_1k_2}]_{k_1,k_2}^K$ denotes the matrix of the nonnegative hyper-parameters, and $\mathbf{Z}=[z_{k_1k_2}]_{k_1,k_2}^K$ is a binary mask. More specifically, we consider the dimensions of the transition kernel as vertices, and the non-zero transition behaviours as graph edges. Naturally, we can capture the sparse structure of the transition kernel $\bm{\Pi}$ using a graph. As shown in Fig.~\ref{fig:2}, for each pair of two vertices $i$ and $j$, $b_{ij}=1$ means that the transition probability from $i$-th component to $j$-th component is non-zero, and vice versa. In particular, we model the binary mask $\mathbf{Z}$ using relational gamma processes as
\begin{align}
    z_{k_1k_2} \sim \mathrm{Ber}[1-\exp(\sum\nolimits_{c=1}^{C} m_{k_1c} r_c m_{k_2c})],\notag
\end{align}
where $r_c$ can be considered as the weight of latent community $c$, and $m_{kc}$ measures how strongly $k$-th vertex (the dimension of the transition kernel) relate to $c$-th latent community, as illustrated in Fig.~\ref{fig:2}. 
Note that the binary mask $\mathbf{Z}$ can be equivalently drawn via the Bernoulli-Poisson link function as
\begin{align}
    z_{k_1k_2} \sim \delta(w_{k_1k_2} \geq 1),\ \ w_{k_1k_2} \sim \mathrm{Pois}(\sum\nolimits_{c=1}^{C} m_{k_1c} r_c m_{k_2c}). \notag
\end{align}
To ensure the model explainability, we restrict $r_c$ and $m_{kc}$ to be nonnegative, and thus place Gamma priors over these two parameters as
$m_{kc} \sim \mathrm{Gam}(\hat{a}_k, \hat{b}_k)$, 
$r_c \sim \mathrm{Gam}(\frac{r_0}{C}, c_0),$ 
respectively. As we discussed in Sec 3.2, this hierarchical gamma prior can be considered as a draw $G=\sum_{c=1}^{\infty}r_c\delta_{\mathbf{m}_c}$ from a gamma process $\mathrm{GaP}(G_0, c_0)$. In particular, we call this Bayesian non-parametric prior the relational gamma process, as a graph-structure can be naturally induced. 
The nonnegative hyper-parameters $\mathbf{D}=[d_{k_1k_2}]_{k_1,k_2}^K$ are drawn from a gamma distribution as 
$d_{k_1k_2}\sim \mathrm{Gam}(\epsilon_0, \epsilon_0)$. 

The proposed gamma dynamical systems are not fully conjugate. Nonetheless, tractable-yet-efficient Gibbs sampling algorithms are developed to perform posterior simulation via negative-binomial data augmentation strategies~\cite{zhou2016nonparametric}. 
The full derivation of the inference procedure is presented in the  supplementary material.

\section{Related Work}
Modeling sequentially observed count sequences has been receiving growing interests in recent years. Here we discuss several types of methods closely related to our studies. 
\cite{acharya2015nonparametric} first studies the gamma Markov process on sequentially count sequences, in which the latent states evolve independently over time. \cite{schein2016poisson} tries to capture the excitations among the latent gamma Markov processes using a transition structure. \cite{schein2019poisson} investigates a Poisson-randomized gamma Markov process which can capture a certain amount of bursty dynamics, and thus demonstrates advantages over gamma Markov processes. \cite{pmlr-v108-virtanen20a} studies a second type of gamma Markov chain structure via the scale parameter of the latent gamma states, which demonstrate better stationary property over the gamma Markov chain proposed by \cite{acharya2015nonparametric}. \cite{2021A} recently provides a thorough survey on the studies of the developed  gamma Markov processes, and evaluates these models through standard tasks including data smoothing and forecasting. 
\cite{han2014dynamic} first tries to capture sequential count observations using linear dynamical systems, via the extend rank likelihood function. \cite{pmlr-v54-linderman17a} proposes to learn switching behaviors of sequential data using recurrent linear dynamical systems (rLDS). \cite{2018Tree} further develops a tree-structured extension of rLDS, with multi scale resolution. \cite{ijcai2020p281} extends the Poisson gamma dynamical systems to learn non-stationary transition dynamics behind count time series. 
Some efforts are also dedicated to developing Bayesian deep models to capture count sequences.~\cite{TSBN} develops a temporal sigmoid belief network for count time series.~\cite{NEURIPS2018_4ffb0d2b} proposes deep Poisson-dynamical systems to capture long-range temporal dependencies.

\section{Experiments}
We evaluate the proposed relational gamma process dynamical systems, and compare it with closely-related methods, using both synthetic and real-world count data.
\begin{figure*}[t]
    \centering
    \includegraphics[scale=0.43]{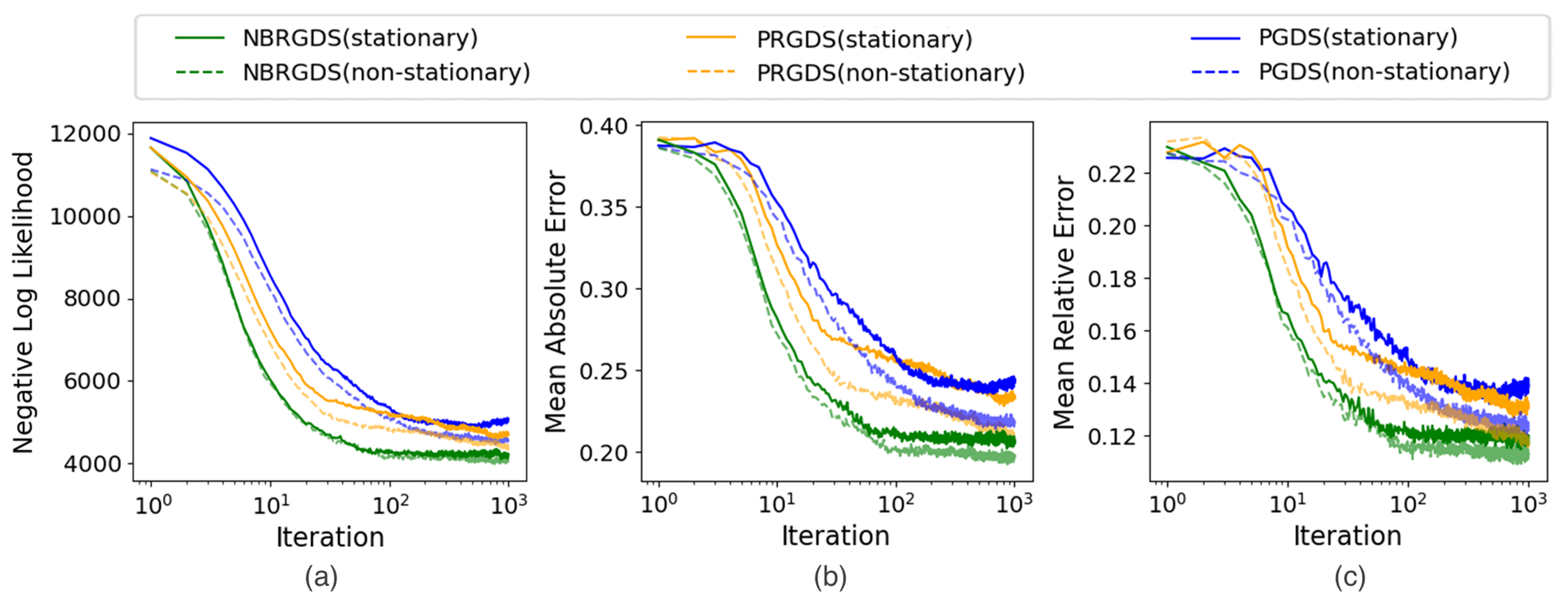}
    \caption{Negative-binomial-randomized gamma dynamical systems (NBRGDSs) demonstrate strong ability in capturing \emph{heterogeneous} overdispersion effects, and thus achieves faster convergence (a), lowest mean absolute error (b) and mean relative error (c), compared with the other related baselines. The stationary and non-stationary generative process i.e. $\delta^{(t)} = \delta$ and $\delta^{(t)}$, denoted as solid line and dotted line, respectively. }
    \label{fig:4}
\end{figure*}

\noindent\textbf{Real-world data. }
We conducted the experiments with the following real-world datasets:  
\textbf{(1) Integrated Crisis Early Warning System (ICEWS)} dataset contains the count number of $6,000$ pairwise interactions between $233$ countries over $365$ days. By screening out $4,800$ dimensions where the sample sparsity exceeds 99\%, we used a subset of ICEW data which contains $V=1,200$ dimensions, and $T=365$ time steps; 
\textbf{(2) Last.fm} contains the listening information of $7,071$ music artists over 51 months, where we have $T=51$ time steps, and $V=7,071$ dimensions;
\textbf{(3) Earthquake Reports Database (EQDB):} records more than $120,000$ earthquake reports  over $15,000$ earthquakes whose epicenters in the United States and nearby U.S. territories from 1928 to 1985. We created a count matrix where each column represents a month and each row represents a state. The EQDB used in the experiments, contains $T=696$ time steps, and $V=64$ dimensions.
\textbf{(4) COVID-19} contains the daily death toll in the fifty states and Washington DC of the United States,  from March 2020 to March 2021. We have $T=365$ time steps, and $V=51$ dimensions.

\noindent\textbf{Baselines.} 
In the experiments we compared the predictive of the proposed models with (1) the gamma process dynamic Poisson factor analysis (GaP-DPFA)~\cite{acharya2015nonparametric}, in which the gamma Markov chain evolves independently over time; (2) the Poisson-gamma dynamical system (PGDS)~\cite{schein2016poisson}, in which a transition kernel is used to capture the excitations among latent gamma Markov chains; (3) the Poisson-randomized gamma dynamical system (PRGDS)~\cite{schein2019poisson}  where the Poisson-randomized gamma Markov chain structure can capture a certain amount of bursty dynamics. 

We denote the proposed negative-binomial-randomized gamma dynamical system as NBRGDS. The proposed NBRGDS with factor-structured prior imposed over the transition kernel, is denoted as FS-NBRGDS. The proposed NBRGDS with graph-structured prior placed over the transition structure, is denoted by GS-NBRGDS.

To evaluate the performance of the compared models in capturing heterogeneous overdispersed behaviours of latent dynamic processes behind count sequences, we considered a subset of ICEWS data that consists of heterogeneous overdispersed counts. More specifically, we sorted the oberserved dimensions according to their variance/expectation ratio, and selected the first $L$ dimensions in descending order, i.e., those dimensions with larger  variation/expectation ratio. We present the result for the setting $L=300$. We treated the 80 percent of the data as the training set, and the remaining 20 percent as the test set. Then, we trained all the compared models with the training set, and evaluated the model performance using the test set. In the experiments, we used mean absolute error (MAE) and mean relative error (MRE) to evaluate the model performance in fitting count sequences: 
\begin{align}
\mathrm{MAE}&=\frac{1}{VT}\sum_{v=1}^V\sum_{t=1}^{T}|n_v^{(t)}-\hat{n}_v^{(t)}|,\notag\\
\mathrm{MRE}&=\frac{1}{VT}\sum_{v=1}^V\sum_{t=1}^{T}\frac{|n_v^{(t)}-\hat{n}_v^{(t)}|}{1+n_v^{(t)}}, \notag 
\end{align}
where $n_v^{(t)}$ and $\hat{n}_v^{(t)}$ denotes the ground true value and estimated value of dimension $v$ at time $t$, respectively. They differ because MRE considers the relative magnitude of the errors in relation to the actual values, taking into account the scale of the data, while MAE simply measures the absolute magnitude of the errors without considering the data scale. Fig.~\ref{fig:4} shows the  results of the compared models averaged over ten random training-testing repeats.
\linespread{1}
\begin{table*}[t]
    \centering
    \small
    \begin{tabular}{llcrrrrrr}
        \toprule
        & & &GaP-DPFA&PGDS&  PRGDS& NBRGDS& FS-NBRGDS&GS-NBRGPDS\\
        \midrule
        \textbf{ICEWS}
        &MAE&S&1.29±0.01&1.04±0.02& 1.06±0.01& \textbf{1.04±0.00}& 1.04±0.01&1.04±0.01\\
        &   &F&0.94±0.01&0.95±0.03& 0.98±0.04& 1.12±0.02& 0.91±0.02&\textbf{0.90±0.01}\\
        &MRE&S&0.61±0.00&\textbf{0.41±0.01}& 0.42±0.00& 0.43±0.01& 0.42±0.00&\textbf{0.41±0.02}\\
        &   &F&0.53±0.01&0.51±0.04& 0.59±0.03& 0.58±0.02& 0.54±0.01&\textbf{0.51±0.01}\\
        \hline
        \textbf{Last.fm}
        &MAE&S&1.71±0.03&1.38±0.01& 1.39±0.01& 1.37±0.01& \textbf{1.36±0.02}&1.38±0.02\\
        &   &F&8.04±0.07& 1.41±0.02& 5.47±0.22& 1.41±0.02& \textbf{1.12±0.01}&1.13±0.02\\
        &MRE&S&0.52±0.01& 0.34±0.01& 0.34±0.00& 0.34±0.01& \textbf{0.33±0.00}&0.34±0.00\\
        &   &F&4.02±0.04& 0.86±0.02&2.59±0.09& 0.85±0.03& 0.53±0.04&\textbf{0.53±0.01}\\
        \hline
        \textbf{EQDB}
        &MAE&S&3.37±0.09&3.37±0.20& 3.41±0.33& \textbf{3.26±0.12}& 3.26±0.20&3.34±0.27\\
        &   &F&10.17±2.21& 10.89±0.94& 7.08±0.90& 5.65±0.63& 3.55±0.04&\textbf{3.33±0.06}\\
        &MRE&S&0.89±0.17& 0.89±0.05& 0.84±0.08& 0.83±0.09& 0.83±0.07&\textbf{0.82±0.05}\\
        &   &F&8.12±2.14&6.54±0.93& 2.45±0.46& 1.49±0.11& 1.52±0.15&\textbf{1.40±0.10}\\
        \hline
        \textbf{COVID-19}
        &MAE&S&12.09±0.26& 11.42±0.62&11.08±0.25& \textbf{10.99±0.43}& 11.57±0.07&11.37±0.62\\
        &   &F&23.35±1.07& 27.67±0.18& 21.95±0.59& \textbf{20.87±0.29}& 23.55±0.08&23.30±0.05\\
        &MRE&S&1.47±0.19& 1.54±0.11& 1.23±0.11& \textbf{1.19±0.13}& 1.41±0.22&1.32±0.11\\
        &   &F&6.30±0.58& 7.87±0.32&6.32±0.32& 1.94±0.15& \textbf{1.36±0.03}&1.40±0.03\\
        \bottomrule
    \end{tabular}
    \caption{Results for the data smoothing (``S'') and future data forecasting (``F'') tasks. For both mean absolute error (MAE) and mean relative error (MRE), lower values are better.}
    \label{tab:1}
\end{table*}


\begin{figure}[t]
    \centering
    \includegraphics[scale=0.33]{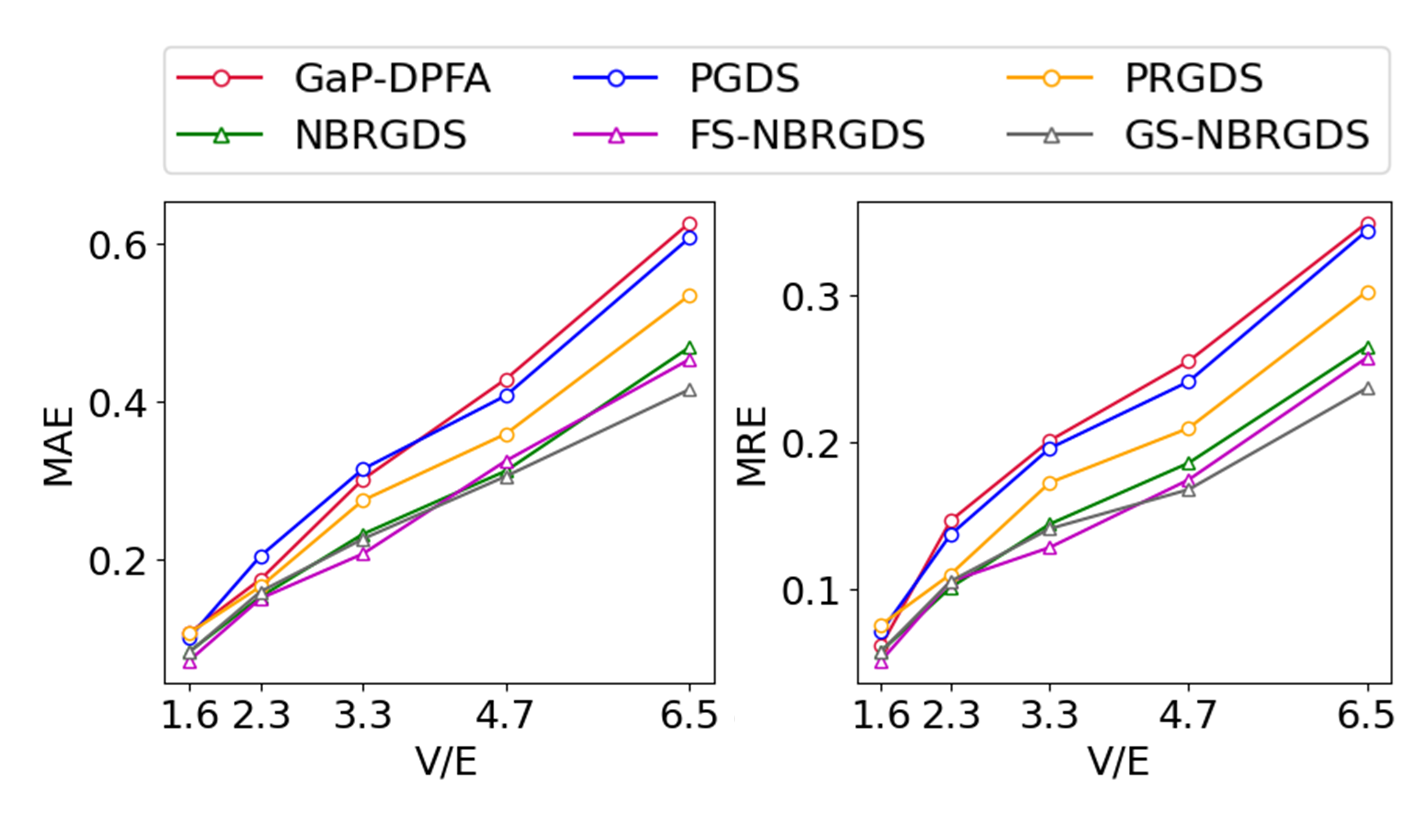}
    \caption{The proposed NBRGDS consistently achieves lower mean absolute and relative errors, when we vary the overdispersed magnitude (the ratio of variance to expection) of the synthetic count sequences, compared with the other closely-related models.}
    \label{fig:5}
\end{figure}
As shown in Fig.~\ref{fig:4} (a), NBRGDS has started to converge to its steady states after almost $10^2$ iterations, while both PGDS and PRGDS start to converge until $10^3$ iterations. Fig.\ref{fig:4} (b) and (c) compares the mean absolute errors and mean relative errors of the compared models, respectively. Overall, NBRGDS achieves the lowest MAE and MRE. PRGDS performs better than PGDS as PRGDS can capture a certain amount of overdispersion effects via its Poisson-randomized chain structure. We also note that NBRGDS with a time-varing scaling factor $\delta^{(t)}$, performs better than stationary NBRGDS because this scaling factor $\delta^{(t)}$ can also capture bursty dynamics. 
Nonetheless,  stationary NBRGDS still outperforms both the PRGDS and PGDS with time-varing $\delta^{(t)}$. We conjecture that this improved prediction accuracy is because the time-varying scaling factor $\delta^{(t)}$ fail to capture the underlying overdispersed behaviours, although it still can model a certain amount of bursty dynamics in observed dimensions. This observation further demonstrates the strong ability of the NBRGDS in capturing hetergeneous overdispersion effects of the latent dimensions behind count sequences.

\begin{figure*}[ht!]
    \centering
    \includegraphics[scale=0.62]{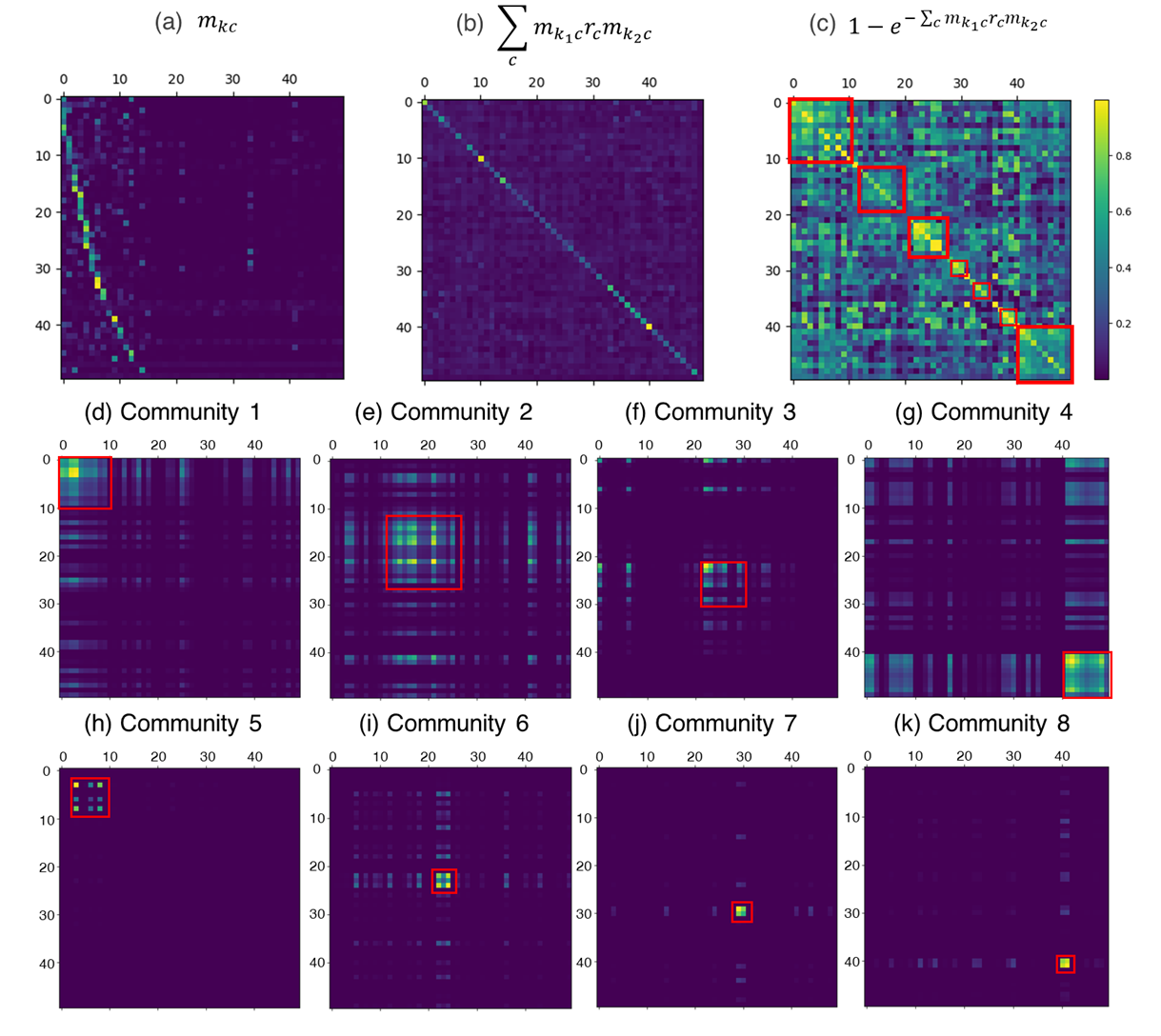}
    \caption{ The latent graph structure inferred by the proposed method on  \textbf{ICEWS} data}
    \label{fig:6}
\end{figure*}

\noindent\textbf{Synthetic data.} 
To further evaluate the performance of the compared models in capturing overdispersion effects, we also considered generating synthetic data with heterogeneous overdispersed dynamics. To that end, we considered to simulate synthetic data using zero-inflated negative-binomial (ZINB) models given by
\begin{align}
    f_{\mathrm{ZINB}}(n \mid p_0, r, p)=p_0I_0(n) + (1-p_0)f_{\mathrm{NB}}(n \mid r, p), \notag
\end{align}
where $f_{\mathrm{ZINB}}$ and $f_{\mathrm{NB}}$ represents the probability mass function (PMF) of the zero-inflated negative-binomial distribution and negative-binomial distribution, respectively. Here, $I_0(n)$ is an indicator function that takes 1 when $n=0$, otherwise 0. The parameter $p_0 \in [0,1]$ controls the ratio of zero counts, while $r$ and $p$ are the two parameters of the negative-binomial distribution. Hence, we can effectively control the sparsity and overdispersion magnitude of the dimensions by tuning the values of $p_0$ and $p$, respectively\footnote{Assume an random variable $x\sim\mathrm{ZINB}(p_0, r,p)$. The expectation and variance of $x$ are $\mathsf{E}[x]=r(1-p_0)(1-p)/p$, and $\mathsf{Var}[x]=(1-p_0)r(1-p)/p^2 + p_0(1-p_0)r^2(1-p)^2/p^2$, respectively. Thus, the ratio of variance to expectation of $x$ is $(1+rp_0(1-p))/p$.}. 
More specifically, we generated five groups of synthetic data, in which each group contains $V=10$ dimensions and $T=365$ time steps, using the following configurations: 
\textbf{(1)} $p_0=0.9$, $r=5$, $p=0.9$, $\mathsf{V}/\mathsf{E} = 1.6$; 
\textbf{(2)} $p_0=0.9$, $r=5$, $p=0.8$, $\mathsf{V}/\mathsf{E} = 2.3$;
\textbf{(3)} $p_0=0.9$, $r=5$, $p=0.7$, $\mathsf{V}/\mathsf{E} = 3.3$;
\textbf{(4)} $p_0=0.9$, $r=5$, $p=0.6$, $\mathsf{V}/\mathsf{E} = 4.7$;
\textbf{(5)} $p_0=0.9$, $r=5$, $p=0.5$, $\mathsf{V}/\mathsf{E} = 6.5$,
where $\mathsf{V}$ and E represents variance and expectation of each group data, respectively. Then $\mathsf{V}/\mathsf{E}$ denotes the ratio of variance to expectation, and thus measures overdispersion effects. 
Fig.\ref{fig:5} plots the model performance of the compared methods by varying the ratio of variance to expectation. NBRGDS models including its factor-structured and graph-structured versions, consistently outperforms the other methods. In particular, NBRGDS achieves a significant improvement compared with PRGDS, although PRGDS still can capture a certain amount of overdispersion effects. Additional experiments on synthetic data under different configurations  are available in the supplementary material.

\noindent\textbf{Data Smoothing and Forecasting}
To quantatively evaluate the predictive performance of the compared methods, we considered two standard tasks: data smoothing and future data forecasting. The data smoothing is to predict $y^{(t)}_v$ given the remaining observations $y^{(1:T)}_v \setminus y^{(t)}_v$, while the future data prediction is to predict future observations at next $S$ time steps $y^{(T+1):(T+S)}_v$ given the history up to $T$, $y^{(1:T)}_v$. Here we considered to predict next two time steps ($S=2$). We used the default settings of GaP-DPFA, and PRGDS as provided in the corresponding releases \cite{acharya2015nonparametric,schein2019poisson}. For NBRGDS, FS-NBRGDS and GS-NBRGDS, we choose $K=100$ when $V\geq1000$, while the dimensions of EQDB and COVID-19 datasets are smaller than $100$, thus we choos $K=25$. We set $C=K$ for FS-NBRGDS and GS-NBRGDS. We ran 5000 iterations of the Gibbs sampler, which have started to converge after 1000 iterations. We discarded the first 3000 samples which were treated as burn-in time and collected a posterior sample every tenth sample thereafter.
Tab.~\ref{tab:1} shows the results of the compared methods in these two tasks. Overall, NBRGDS outperforms the GaP-DPFA, PGDS and PRGDS on almost all the datasets. In particular, we found that FS-NBRGDS and GS-NBRGDS show superior performance in future data forecasting. We conjecture this improved prediction accuracy is due to that FS-NBRGDS and GS-NBRGDS can effectively leverage the structure information underlying dynamic count data, and thus yields better predictive accuracy.
We provide more comparative results on data smoothing and forecasting over different models in appendix Sec.D.

\noindent\textbf{Graph-Structured Transition Dynamics.} 
Fig. \ref{fig:6} shows the latent graph structure underlying the transition kernel, estimated by the proposed model. Although the model was initialized with $C=50$ latent communities, the latent graph only consists of approximately ten communities with non-zero weights, as shown in Fig. \ref{fig:6}(a). Fig. \ref{fig:6}(d-k) plots the eight evident latent communities in which the vertices are densely connected as the corresponding dimensions are more likely to interact with each other. Fig. \ref{fig:6}(c) demonstrates that the most estimated latent dynamic processes are almost independent to the other dynamic processes, but only interact with a few other dimensions. We provide the latent graphs inferred for the other real-world data in the supplement.
\section{Conclusion}
Novel negative-binomial-randomized gamma dynamical systems, have been proposed to capture the \emph{heterogeneous} overdispersed behaviors of latent dynamics behind count time sequences. 
The new framework demonstrates more explainable latent structure, by learning the factor structure and sparse graph structure of the transition kernels, compared with transition kernel by non-informative priors. Although the prior specification of the proposed framework lacks conjugacy, tractable-yet-efficient sampling algorithms are developed to perform posterior inference.  
In the future, we plan to capture time-varying graph-structured transition dynamics, which will enable to better understand non-stationary count sequences. We are also considering to enhance the modeling capacities of gamma belief networks \cite{zhou2015poisson,JMLR:v17:15-633,NEURIPS2018_efb76cff} and convex polytope methods~\cite{zhou2016softplus,CVXT}  using the negative-binomial-randomized gamma Markov processes. Moreover, the future interesting research directions include modeling \emph{bursty} dynamics often observed in online or mobile social networks~\cite{AAAI-18,ICML-18,UAI-20,SDM-23,AAAI-24} in the future research.

\bibliographystyle{named}
\bibliography{ijcai24}
\end{document}